# Early Prediction of 30-day ICU Re-admissions Using Natural Language Processing and Machine Learning

**Zhiheng Li**
Northeast Yucai Foreign Language School
Shenyang, Liaoning, China
Email: 1479551965@qq.com

**Xinyue Xing**
Northeast Yucai Foreign Language School
Shenyang, Liaoning, China
Email: 1045068673@qq.com

**Bingzhang Lu**
Northeast Yucai School
Shenyang, Liaoning China
Email: 2268665795@qq.com

**Zhixiang Li\***
Department of Biomedical Engineering
Shenyang Pharmaceutical University
Liaoning, China
Email: 106040205@syphu.edu.cn

**Abstract:**

ICU readmission is associated with longer hospitalization, mortality and adverse outcomes. An early recognition of ICU re-admission can help prevent patients from worse situation and lower treatment cost. As the abundance of Electronics Health Records (EHR), it is popular to design clinical decision tools with machine learning technique manipulating on healthcare large scale data. We designed data-driven predictive models to estimate the risk of ICU readmission. The discharge summary of each hospital admission was carefully represented by natural language processing techniques. Unified Medical Language System (UMLS) was further used to standardize inconsistency of discharge summaries. 5 machine learning classifiers were adopted to construct predictive models. The best configuration yielded a competitive AUC of 0.748. Our work suggests that natural language processing of discharge summaries is capable to send clinicians warning of unplanned 30-day readmission upon discharge.

# Contents

## 1. Introduction:

ICU readmission describes the situation where patients get readmitted into the ICU within a relatively short interval (30 days as recognized in most cases). Such readmission is usually accompanied by deteriorated symptoms and thus longer ICU stay as well as higher mortality rate. Predicting ICU readmission can help us recognize potential causes of such unwanted situation. The most effective therapy will be concluded accordingly so as to prevent readmissions and the problems that may follow.

EHR (Electronic Health Record) stands for an electronically stored data base containing health conditions of a broad range of population. To be specific, the conditions include "demographics, medical history, medication and allergies, immunization status, laboratory test results, radiology images, vital signs, personal statistics like age and weight, and billing information"[1]. Due to its digital nature, EHR can be shared among institutions conveniently. Its abundance in information also makes it a perfect source in medical decision making.

Machine learning has long been a popular measure of constructing prognostic models on the basis of abundant data. While unsupervised learning generates unified categorizations out of irregular data patterns, supervised learning trains the existing data to establish prophetic models and predict potential instances accordingly. Given the collaborative ability of machine learning, and that the EHRs cannot be easily integrated manually, machine learning is widely used to process EHRs and further contribute a great deal in medical decision making. To this end, we consider to associate the plentiful resources of EHR and the mighty models of machine learning to construct comprehensive analytical models and therefore produce reliable predictions of ICU readmissions.

## 2. Related Works:

It is natural that numerous sets of data need to be processed during pharmaceutical researches and developments so as to ensure the safety of patient consumption. Traditionally, the process generally involves medical information received from patients, healthcare providers, medical literature, physicians, pharmaceutical company's sales team, pharmacists, or the like. The pharmacovigilance system, notably, is a system capable of complete end-to-end processing of adverse event (AE) reaction reports that addresses problems of manual errors and inconsistent report generations. Nevertheless, the system's reliance on the large amount of diverse data posts strict limitations to its applications. [2]

Established originally for observational studies, EHR has now earned an irreplaceable position in the medical research world due to its abundance in easily accessible and comprehensible medical records. While most analytical systems applied by researchers require loads of data to produce desired results, and the data could only be arduously collected from different organizations, EHRs offer profuse collaborated data and thus provide us opportunities to enhance patient care, embed performance measures in clinical practice, and improve the identification and recruitment of eligible patients and healthcare providers in clinical research. Observational studies, safety surveillances and clinical researches have all made abundant uses of EHRs, under the presence of which the labor-intensive data collecting tasks are about to become history.[3]

NLP-based computational phenotyping has a broad range of applications. However, it's never easy to construct such computational phenotyping manually. Among the by-far most productive methods of constructing, which are keyword search and rule-based system, tons of work need to be done to produce the list of keywords or rules. Supervised machine learning models, in this case, are extremely favored due to their capability of recognizing and extracting properties useful for producing proper classifications out of the irregular data. Deep learning and unsupervised learning are also applied to discover novel phenotypes. With these machine learning patterns combined, data can be integrated to offer the simplest approach possible for further applications [4].

This research synthetically relates to our work. While life expectancy is a major factor that influences decisions about one's final phase of life, proper prognostication can help minimize cost of resources and maximize efficiency of treatments. Given the irregular and unified descriptions included in EHRs, the researchers applied machine learning and language processing to analyze loads of data in search for models of interrelationships between a person's life expectancy and past health conditions. Various models have been tested and the best fit has been found to produce the most accurate prediction of one's life expectancy. The research proved that the combination of machine learning and natural language processing offers feasible pathways to making reliable predictions over complicated cases.[5]

**3. Methods:**

Detailed steps of how we built the 30-day unplanned ICU readmission predictive model can be found in Fig.1. We will introduce it step by step in the following subsections.

**3.1 Dataset:**

Data for this study was a subset extracted from the public accessible critical care database, named MIMIC-III [6]. This dataset contains over 40,000 patients' de-identified data when the patients were in ICU unit. We first extracted all patients' admission data from the table named ADMISSIONS. Next, we excluded patients with the admission type of "NEWBORN", as the newborn patients' information might be archived in other database. The part of patients also has a high missing rate of discharge summaries. Then, a hospital-expiration-flag, which describes whether patients pass away during hospitalization, was used to discard those patients must not have re-admission due to mortality. Then, if one ICU stay has a straight following "ELECTIVE" admission, we would seek the next "non-ELECTIVE" admission to be the re-admission, as we focused on unplanned readmission. We should notice that patients with multiple ICU stays were treated as independent subjects. We used subject-id and row-id to link admission table with the notes table. The discharge summary of each selected subject was then found. Some odd cases, such has multiple discharge summaries of one stay, were then excluded. Finally, a cohort of 45305 subjects were generated, among which 5.26% has unplanned readmission. The distribution of readmission time interval can be found in Fig.2.

**3.2 Word and Concept representation:**

The discharge summaries were first pre-processed with Porter stemming to reduce inflectional variations. In order to downstream to the machine learning classifiers, the raw text of summaries should be converted to vectors. We employed Bag-of-words (BoW) as the vectorization techniques for whose simply implementation and great success in language modeling and document classification. With "BoW", each word in the summaries can be represented into a value while one summary can be represented into a vector. As "BoW" aims at extract important words from one document to distinguish this document from others with the help of these important words, some exclusion criteria should be applied to improve the performance and highlight important words. We excluded 313 stop words as the reference from NCBI. The words occurred in more than 95% summaries were also discarded. To avoid that excessive features would bring difficulty to the machine learning classifiers, words occurred less than 5 summaries were also excluded. Term frequency-inverse document frequency (tf-idf) weighting adjustment was also applied. Finally, 22,405 words were included to form the bag. In other words, each discharge summary was represented to a 22,405-long vector.

Considering about the variation and written habits when generating the discharge summaries, there was plenty of inconsistency to claim a same clinical meaning. This brought great difficulties to over model as BoW only care about the words. To address this problem, we employed the phenotyping system named MetaMap server from Unified Medical Language System (UMLS) to identify medical concepts from the discharge summaries. Each concept has a unique identifier named Concept Unique Identifiers (CUIs). We also generated Bag-of-CUIs as the feature representation from discharge summaries.

Briefly, two feature sets: Bag-of-Words and Bag-of-CUIs were generated by interpreting discharge summaries using NLP techniques. These 2 feature sets would be down-streamed to the machine learning classifiers.

**3.3 Machine Learning Classifiers and Pipeline:**

5 machine classifiers were selected to build the predictive models, including 2 linear classifiers: logistic regression and support vector machine with linear kernel (LinearSVM), 2 ensemble models: random forest and gradient boosting decision trees (GBDT), and Naive Bayes as the benchmark. Entire dataset was first stratified to training set: held-out test set as 7:3. There were 5 algorithms on 2 feature sets, which were totally 10 configurations. We adopted area under the receiver operating characteristic (AUC) to evaluate the performance as the binary classification is highly imbalanced, Parameters of each algorithms were tuned with 5-fold cross-validation on the training set and then examined on the held-out test set.

**3.4 Tools:**

We built our configurations over Python 3.6.3. Machine learning classifiers, cross-validation, data pre-processing, parameters tuning was implemented by Scikit Learn package. The experiments were deployed on a Google Could Server, which has 8 CPUs and is capable of tuning parameters and conducting cross-validation in a timely manner.

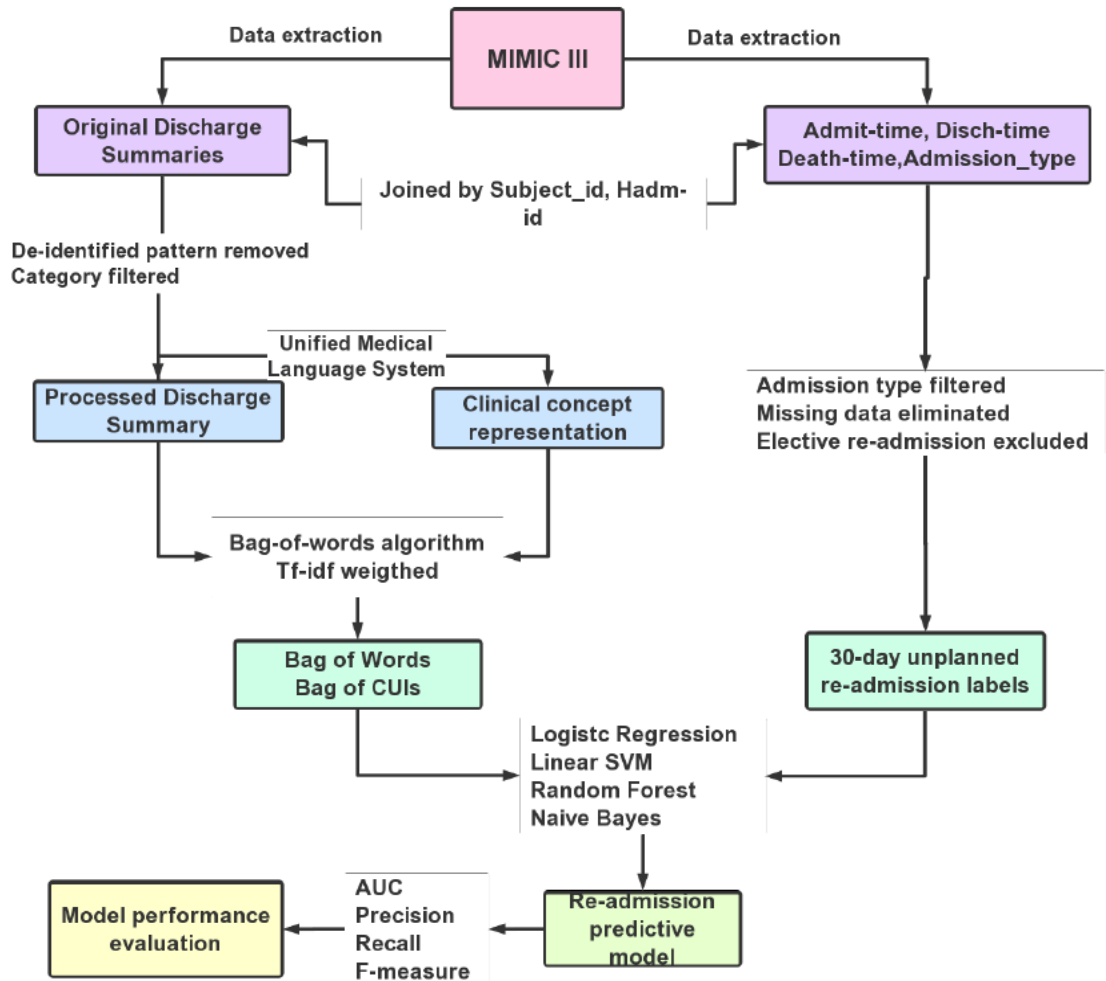

**Figure 1**: The study design. We extracted patients' discharge summaries from MIMIC III database. Exclusion and inclusion criteria were applied to extract a cohort of patients with label of whether has a 30-day unplanned hospital readmission. We integrated 2 different data representation methods and 6 machine learning algorithms to generate predictive models. Area under receiver operating characteristic curve (AUC) was used to evaluate the model performance.

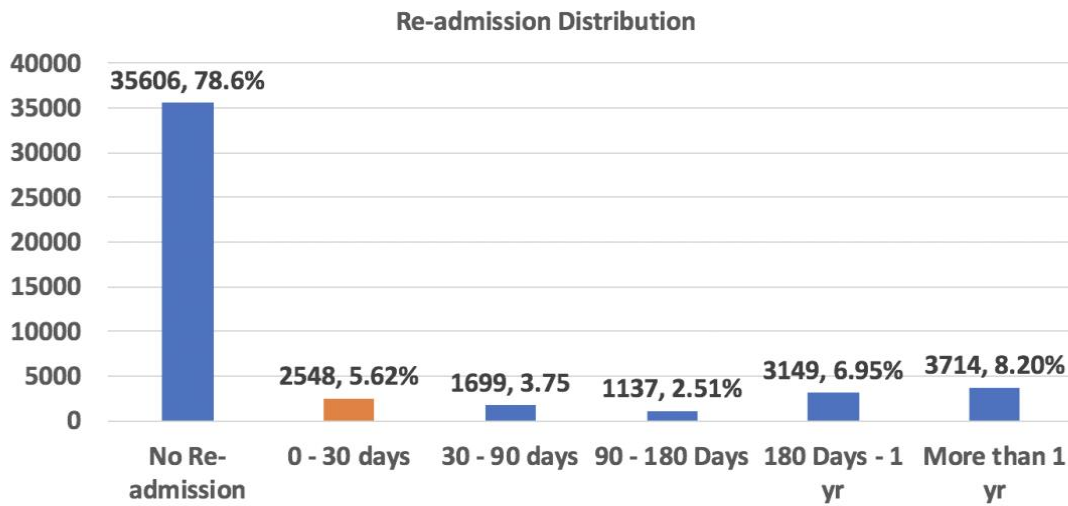

**Figure 2**: Readmission distribution. There were 45305 subjects in the cohort, among which 2548, 5.62% subjects have 30-day unplanned ICU readmission.

**4. Results and Analysis:**

**4.1 Results:**

In this part, we will report the evaluation results of 5 supervised learning classifiers over 2 feature sets. From Tab.1, we can see that except Naive Bayes as the benchmark, all other 4 classifiers yielded an AUC over 0.7. Logistic regression over Bag-of-CUIs had the best performance of AUC: 0.748, followed by LinearSVM over the same feature yielded AUC: 0.744. The best performance of Bag-of-Words was 0.743, which is the 3rd place among all 10 configurations.

We further compared the performance of 2 features [7]. Bag-of-CUIs beat BoW on 2 linear classifiers while lost on 2 ensemble classifiers. This might because that each word or phrase was mapped to multiple medical concepts in Metamap, which would also result in more features in Bag-of-CUIs than BoW. According to the experience of previous experiment, linear classifiers were better than ensemble classifiers at handling larger scale feature space. The performance of this study also validated that logistic regression and LinearSVM were the best classifiers when integrating Bag-of-words and machine learning to build clinical predictive models.

**4.2 Analysis:**

We further examined our models by presenting the top 20 features with greatest feature importance in logistic regression model over both BoW and Bag-of-CUIs as Fig. 3. The font size of each feature is associated with its importance towards the models. Most of the words in both feature space were clinical meaningful. In Bag-of-Words, we could find "Tracheotomy" (which is a surgical procedure which consists of making an incision on the anterior aspect of the neck and opening a direct airway through an incision in the trachea) and "fistula" (an abnormal connection between organs). In Bag-of-CUIs, we could find "CORONARY-ARTERY-BYPASS-SURGERY" (which is a surgical procedure to restore normal blood flow to an obstructed coronary artery) and "SUBDURAL-HEMATOMA" (A pool of blood between the brain and its outermost covering). All these words and concepts are capable to result in unplanned readmission due to severe condition or post-surgical morbidity. We also found some features emerged in both feature space. Such as "bipap" in BoW is the abbreviation of "BILEVEL-POSITIVE-AIRWAY-PRESSURE" in Bag-of-CUIs; and "fistula" in BoW is the second word of "ARTERIOVENOUS-FISTULA" in Bag-of-CUIs. These words and concepts presented high correlation with unplanned ICU readmission, can also remind clinicians that these subgroup of patients have a high possibility of readmission. Internal consistency of two feature sets also suggested that clinical meaningful keywords extracted from discharge summaries can build unplanned readmission predictive models and bag-of-words model was able to extract those keywords from discharge summaries.

|  | Bag-of-Words | Bag-of-CUIs |
|---|---|---|
| Naive Bayes | 0.612 | 0.591 |
| LinearSVM | 0.743 | 0.744 |
| Logistic Regression | 0.739 | **0.748** |
| Random Forest | 0.713 | 0.688 |
| Gradient Boosting Decision Trees | 0.721 | 0.704 |

**Table 1**: The performance of 5 machine learning classifiers on ICU readmission prediction using 2 different features as input.

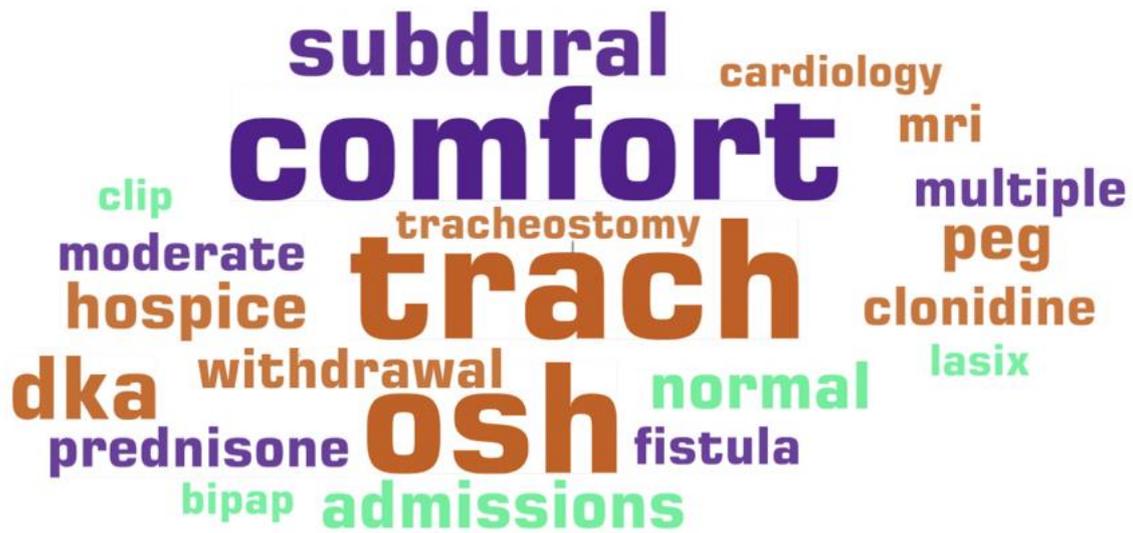

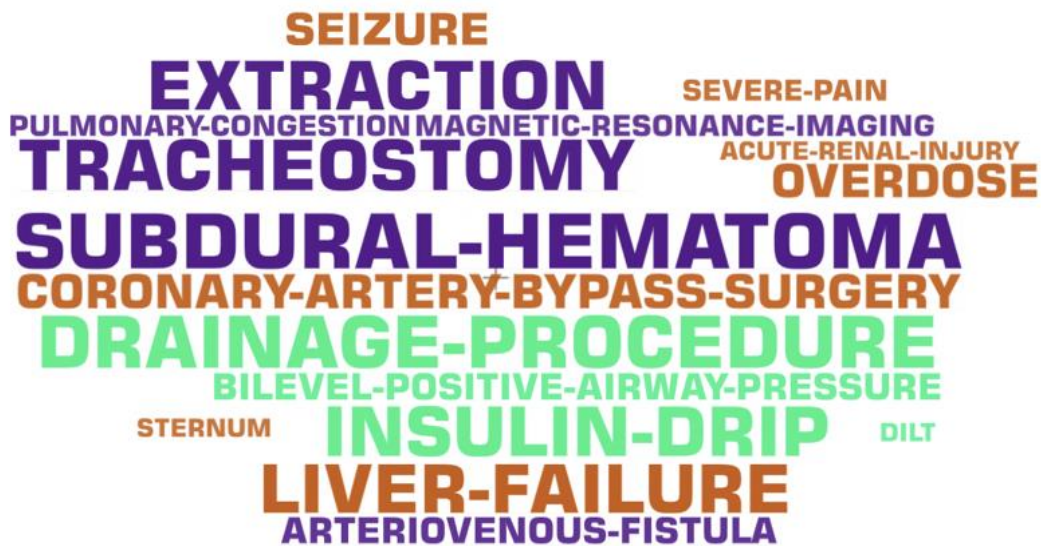

**Figure 3**: Top 20 Words and CUIs have the greatest feature importance in the predictive model. Font size of each feature is associated with the importance. Upper: Bag-of-Words, Lower: Bag-of-CUIs

**4.3 Limitation and Future Work:**

Obviously, there are limitations of out study that could be improved in the future work. Firstly, we only applied traditional machine learning classifiers other than deep neural networks. Recently, deep learning has shown its promising abilities in medical domain. We could apply 1-D convolutional neural network or recurrent neural network in the future experiment. Secondly, our dataset was limited. We only involved subjects from MIMICIII dataset. Future work on more generalized dataset could be done to make our predictive model more solid and robust. Thirdly, we only took discharge summaries into consideration.

However, other clinical notes, such as radiology reports, physicians' diagnosis are also very essential to describe patients' condition within ICU stay. We are planning to involve more clinical notes under other categories. Finally, we only adopted Metamap as the phenotyping system to identify medical concepts from free text. Further experiments could be done with other system such as cTAKE, CLAMP to examine whether Metamap is the most ideal tool in readmission prediction with discharge summaries.

**5. Conclusion:**

Our work validates that carefully representing of discharge summaries is useful to estimate 30-day unplanned ICU readmission. We showed that mapping discharge summaries to unified medical concepts with Metamap as features were capable to yield a competitive AUC of 0.748 when using Logistic Regression as classifier. Our model only needs discharge summaries as the input while previous studies rely on extra lab or radiology results. Future work could be done on enlarging dataset and enriching algorithms.

**References:**

[1] "Mobile Tech Contributions to Healthcare and Patient Experience". *Top Mobile Trends*. 22 May 2014. Archived from the original on 30 May 2014. Retrieved 29 May 2014.

[2] Maitra, A., Mohamedrasheed, A.K., Jain, T.G., Shivaram, M., Sengupta, S., Ramnani, R.R., Pathak, N., Banerjee, D. and Sahu, V., Accenture Global Services Ltd, 2016. "System for automated analysis of clinical text for pharmacovigilance". *U.S. Patent Application* 14/826,575.

[3] Cowie, M.R., Blomster, J.I., Curtis, L.H., Duclaux, S., Ford, I., Fritz, F., Goldman, S., Janmohamed, S., Kreuzer, J., Leenay, M. and Michel, A., 2017. "Electronic health records to facilitate clinical research". *Clinical Research in Cardiology*, 106(1), pp.1-9.

[4] Zeng, Z., Deng, Y., Li, X., Naumann, T. and Luo, Y., 2018. "Natural language processing for EHR-based computational phenotyping". *IEEE/ACM transactions on computational biology and bioinformatics*, 16(1), pp.139-153.

[5] Beeksma, M., Verberne, S., van den Bosch, A., Das, E., Hendrickx, I. and Groenewoud, S., 2019. "Predicting life expectancy with a long short-term memory recurrent neural network using electronic medical records". *BMC medical informatics and decision making*, 19(1), p.36.

[6] UMLS: UCI Machine Learning Repository
https://www.nlm.nih.gov/research/umls/index.html

[7]  "Firefox." en.wikipedia.org. https://en.wikipedia.org/wiki/Firefox (accessed September 13, 2019).